\documentclass[runningheads]{llncs}
\usepackage{graphicx}
\usepackage{comment}
\usepackage{amsmath,amssymb} % define this before the line numbering.
\usepackage{color}

\usepackage{adjustbox}

\newcommand{\Tref}[1]{Table~\ref{#1}}

\newcommand{\fref}[1]{Fig.~\ref{#1}}

\begin{document}
% \renewcommand\thelinenumber{\color[rgb]{0.2,0.5,0.8}\normalfont\sffamily\scriptsize\arabic{linenumber}\color[rgb]{0,0,0}}
% \renewcommand\makeLineNumber {\hss\thelinenumber\ \hspace{6mm} \rlap{\hskip\textwidth\ \hspace{6.5mm}\thelinenumber}}
% \linenumbers
\pagestyle{headings}
\mainmatter
\def\ECCVSubNumber{4585}  % Insert your submission number here

\title{Cross-Identity Motion Transfer \\for Arbitrary Objects through \\ Pose-Attentive Video Reassembling}

% INITIAL SUBMISSION 
\begin{comment}
% \titlerunning{ECCV-20 submission ID \ECCVSubNumber} 
% \authorrunning{ECCV-20 submission ID \ECCVSubNumber} 
% \author{Anonymous ECCV submission}
% \institute{Paper ID \ECCVSubNumber}
\end{comment}
%******************

% CAMERA READY SUBMISSION
% \begin{comment}
\titlerunning{Cross-Identity Motion Transfer}
% If the paper title is too long for the running head, you can set
% an abbreviated paper title here
%
\author{Subin Jeon\inst{1}\orcidID{0000-0003-1651-2249} \and 
Seonghyeon Nam\inst{1}\orcidID{0000-0003-1266-3568} \and 
Seoung Wug Oh\inst{1}\orcidID{0000-0002-8498-0864}  \and \\ 
Seon Joo Kim\inst{1,2}\orcidID{0000-0001-8512-216X}}
\authorrunning{Jeon, Nam, Oh, and Kim}
% First names are abbreviated in the running head.
% If there are more than two authors, 'et al.' is used.
%
\institute{Yonsei University\\\email{\{subinjeon, shnnam, sw.oh, seonjookim\}@yonsei.ac.kr}\and Facebook}
% \emailP
% \end{comment}
%******************
\maketitle
\begin{abstract}
We propose an attention-based networks for transferring motions between arbitrary objects.
Given a source image(s) and a driving video, our networks animate the subject in the source images according to the motion in the driving video.
In our attention mechanism, dense similarities between the learned keypoints in the source and the driving images are computed in order to retrieve the appearance information from the source images.  
Taking a different approach from the well-studied warping based models, our attention-based model has several advantages.
By reassembling non-locally searched pieces from the source contents, our approach can produce more realistic outputs.
Furthermore, our system can make use of multiple observations of the source appearance (e.g. front and sides of faces) to make the results more accurate.
To reduce the training-testing discrepancy of the self-supervised learning, a novel cross-identity training scheme is additionally introduced.
With the training scheme, our networks is trained to transfer motions between different subjects, as in the real testing scenario.  
Experimental results validate that our method produces visually pleasing results in various object domains, showing better performances compared to previous works.
%We validate our method against other motion transfer methods on various object domains
%We validate that our method produces visually pleasing results in various object domains 
% We also provide extensive ablation and subjective user studies to analyze and evaluate our framework.

\keywords{Motion transfer, Generative Adversarial Network, Video to video translation}

\end{abstract}
\begin{figure}[t]
\begin{center}
\includegraphics[width=1\linewidth]{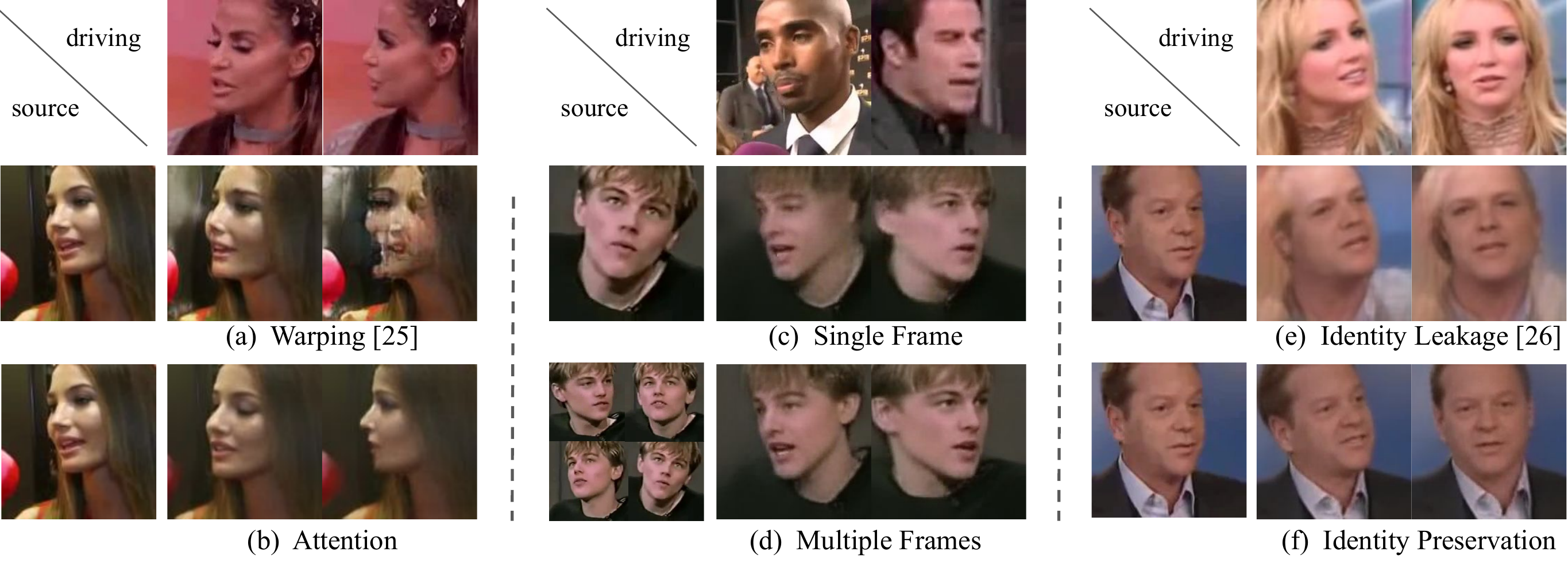}
\end{center}
  \caption{(a) Previous motion transfer methods is based on spatial warping of single image. (b) We propose a non-local search based approach that handles a wide range of motion. (c) A single source image contains insufficient appearance data. (d) Our method can utilize multiple sources with various views. (e) Self-supervision causes identity-leakage in motion transfer. (f) We propose a cross-identity training which is effective in preserving identities of a source image.  }
\label{fig:teaser}
\end{figure}

\section{Introduction}
Motion transfer is a task of transferring motion between different subjects. 
In other words, it generates a video conditioned on the appearance extracted from source image(s) and the motion patterns from a driving video. 
Here, the appearance means an identity of a subject observed in source image(s) and the motion patterns refer to a sequence of poses that change continuously.
Motion transfer has long been studied for its practical applications such as video editing and virtual/augmented reality.
% In this area, researches have been conducted in the direction of designing object-specific models, especially the face models~\cite{blanz1999morphable}.
Recent progress in deep generative model has shown the capability of generating highly realistic images, and it has led to vigorous attempts to transfer motions in learning basis~\cite{Wang:2018:vid2vid,Bansal:2018:RecycleGAN,chan:2019:everybody,Wiles:2018:X2Face,Siarohin:2018:monkeynet,wang:2019:few-vid2vid}. 

%기존 방법
The majority of previous learning-based approaches relies on pre-computed (or given) landmark annotations to represent poses~\cite{Wang:2018:vid2vid,chan:2019:everybody,ma:2017:PoseGuided,tran:2017:disentangledface}.
With an accurate pose representation, those methods have achieved impressive results especially in the human face and body domains where the keypoint extraction is well-studied.
However, the dependency on pre-computed keypoints limits their application.
%at the same time. 
For example, \cite{ma:2017:PoseGuided,tran:2017:disentangledface} are only applicable to specific domains (e.g. face and body), and  \cite{Wang:2018:vid2vid,chan:2019:everybody} work only for specific subjects.

To overcome the above issue, several attempts have been made to transfer object motions without using pre-defined pose representations~\cite{Siarohin:2018:monkeynet,Wiles:2018:X2Face,siarohin:2019:Taichi} recently. 
They loosened the constraint on requiring the pre-defined keypoints by employing the self-supervised learning scheme for the pose representation~\cite{denton:2017:disentangledVideo}, enabling applications beyond specific object categories. 
To learn to extract representations and transfer motion simultaneously, previous methods commonly take warping-based approach to synthesize images in various poses from a source image.
However, we argue that the warping operation is not the best option for this task due to several limitations.   
In this paper, we highlight those limitations and propose a novel attention-based approach for arbitrary motion transfer.

While the warping operation works well for handling small motions, it has some challenges in achieving realistic synthesis.
First, it is difficult to model large and complex motions.
For example, it is more difficult to create a side view than to raise head slightly from the source image with a front view, as illustrated in~\fref{fig:teaser} (a).
Second, it is hard to make use of more than one source image. 
Most of previous works cannot use more than one source image by its design~\cite{ma:2017:PoseGuided,tran:2017:disentangledface,Siarohin:2018:monkeynet}, and some methods rely on heuristics to combine results obtained with several source images~\cite{Wiles:2018:X2Face}.
Observations from different poses and views are essential in completely recovering occluded parts in various views.
As shown in~\fref{fig:teaser} (d), using multiple sources is beneficial in synthesizing various views. 

% The amount of appearance information that a single source image contains is too limited to cover various views.
% Observations from different poses and views are essential to completely recover occluded parts. 
% As different poses under different view require observation from different view and regions previously occluded. 
% However, the amount of appearance information that single source image can contain is too limited to cover various views and occluded parts. 

% 우리 방법 motivation % 우리 방법 작동 방식(Non-local)
We propose a novel attention-based method to transfer a variety of motion from arbitrary object categories.
The main idea of our method is to reassemble the source image by finding proper visual appearance in the source image for each part of the driving pose.
Compared with the warping-based approaches, our method produces more realistic output as it non-locally searches the best source content to synthesize every parts of an image.

% % The intuitive idea of our method is that 
% Our network generates output based on source appearance in a pose similar to the driving pose. 
% Specifically, it first makes separate representations for pose and appearance from the source images. 
% To retrieve appearance similar to driving pose, it compares source poses with driving pose in non-local manner, resulting in robustness to drastic change of pose between source and driving images. %~\fref{fig:teaser} (b)

% 여러 프레임 사용 관점
Furthermore, our framework is flexible in the number of source images. 
While our approach already shows better results than the previous approaches with a single source image, we can maximize the performance by providing additional observations.
Taking the advantage of multiple source images, we can potentially resolve the occlusion issue in the single observation case. 
Previous warping-based methods are designed only to use single source image and they focus on synthesizing unseen parts or different views with the limited appearance information.
We let our network leverage multiple source images by broadening the comparison region to multiple source images and it shows the capability to utilize appearance in various views and poses (\fref{fig:teaser}(c),(d)).

% Cycle reconstruction 
In addition, we propose a cross-identity training scheme that enables realistic motion transfer between subjects with significantly different appearances.
Previously, motion transfer networks are usually trained by transferring poses in the same subject even though the actual task is to transfer poses between different subjects. 
While this training trick is required to learn the networks in an unsupervised way, it is not appropriate for the real testing scenario.
For example, models trained in this way often fails to maintain the personal shape that is unique for each subject independent from its poses (\fref{fig:teaser}(e)).
With the proposed training scheme, the network can learn to perform the motion transfer between different subjects, as in the real testing, without additional supervision. 
Furthermore, our cross-identity reconstruction loss encourages the network to preserve the identity-specific property (e.g. the shape of a human face) after the motion transfer. 

% In other words, it helps our networks to learn more abstract and higher level pose representation.
% In other words, our method encourages the network to disentangle the pose and the identity-specific property.

% dataset, evaluation, user-study
% Extensive experiments demonstrate the efficacy of our method in transferring motion even under drastic changes in pose. We test our network on various objects.% representative deformable objects, human face and body. 
% Moreover, our network shows superior image quality in various views by exploiting multiple source images. 

% We demonstrate our efficacy of transferring motion by conducting quantitative and qualitative experiments on VoxCeleb2~\cite{Chung:2018:VoxCeleb2} and taichi datasets. Our network shows robustness to a wide range of motion compared with previous warping-based approaches. Furthermore, our network shows improvement when exploiting multiple source images.   

% \snam{Please summarize our contribution below.}
To sum up, our contribution can be summarized as follows:
\begin{itemize}
    \item We propose a pose-attentive video reassembling network for motion transfer of arbitrary objects. Our network implicitly learns the pose representation specialized for transferring motion without annotation.
    Through the non-local search based reassembling of visual pieces, we overcome the limitation of warping based motion transfers.
    \item Our framework can naturally process and 
    take advantage of multiple images or video as the source, which helps to synthesize more realistic outputs.
    \item We propose a cross-identity training method to preserve identity-relevant properties such as object shape while changing the pose.% in the case of cross-identity motion transfer. 
    This training strategy matches the testing scenarios, therefore, is more realistic compared to previous works that train on same subjects.
\end{itemize}
\section{Related work}
Remarkable achievements in the generative modelling~\cite{Goodfellow:2014:GAN} have given rise to interests in video generation, leading to extensive development of video generation methods~\cite{vondrick:2016:VGAN,saito:2017:TGAN,Tulyakov:2018:MoCoGAN,villegas:2017:prediction,zhao:2018:motion_forecasting,denton:2018:stochastic_prediction}. 
As a type of video generation, motion transfer refers to the task of synthesizing motion extracted from a driving video on different subjects.
In this section, we address major approaches of motion transfer.

\noindent\textbf{Video-to-video Translation}
Motion transfer has been addressed from the perspective of video-to-video translation.
This class of works ~\cite{Wang:2018:vid2vid,Bansal:2018:RecycleGAN,chan:2019:everybody} proposed methods to learn a mapping between two videos in order to convert a domain of videos into another.
In those frameworks, the generator memorizes diverse aspects of the source appearance using a sequence of frames in the source video and learns to transfer the appearance on the driving video. 
These methods have the advantage of utilizing the rich appearance information of diverse pose and view in the source video, making realistic results.
However, they require time-consuming training per each subject using a large amount of source images related to the subject to train the subject-specific generator.

\noindent\textbf{Pose Guided Image Generation} 
This class refers to the image generation conditioned on a source image and a driving pose. 
In order to represent the pose to condition on, they use keypoint annotations or off-the-shelf keypoint detector. 
Using the keypoint annotations and a single source image, many works ~\cite{ma:2017:PoseGuided,ma:2018:desentangledPerson,siarohin:2018:deformableHuman,balakrishnan:2018:unseen,kulkarni:2015:invgraphics,tran:2017:disentangledface,ding:2018:exprgan} attempted to generate the source subject in a novel pose by warping a single source image.
This approach has strength in transferring motion on new subjects at the test time, but experiences difficulties caused by insufficient appearance information in the single source image.
Recently, multi-source based methods~\cite{zakharov:2019:fews,wang:2019:few-vid2vid,lathuiliere:2020:attention,Ha:2020:marionnette} have emerged to overcome the drawback of a single source image.
These works are similar to our work in that they can transfer motion on new subjects with multiple source images. 
However, we tackle more challenging case because we need to extract pose from images directly and generalize to various object categories, whereas this stream of works is restricted to a specific object category (human body or face).

\noindent\textbf{Motion transfer for arbitrary objects}
This class of works learns to extract pose representation implicitly from driving frames to transfer motion, extending object scopes to arbitrary objects.
X2Face~\cite{Wiles:2018:X2Face} and Monkey-net~\cite{Siarohin:2018:monkeynet} transfer motion by learning to warp a source image to pose of a driving image in unsupervised way.
These warping-based method showed superiority in transferring local motion, but experienced difficulties in synthesizing motion when the movement becomes larger and more complex.
To solve the large motion problem, Siarohin~\cite{siarohin:2019:Taichi} et al. enhanced Monkey-net by taking the local movement into consideration using local affine transformation of keypoints.
Our method also falls in to this category  of transferring motion on arbitrary objects. 
Different from previous works, we propose a non-local motion matching mechanism to capture drastic motion changes in order to  overcome the limitation of  previous warping-based approaches.

\section{Method}

\begin{figure}
    \centering
    \includegraphics[width=\linewidth]{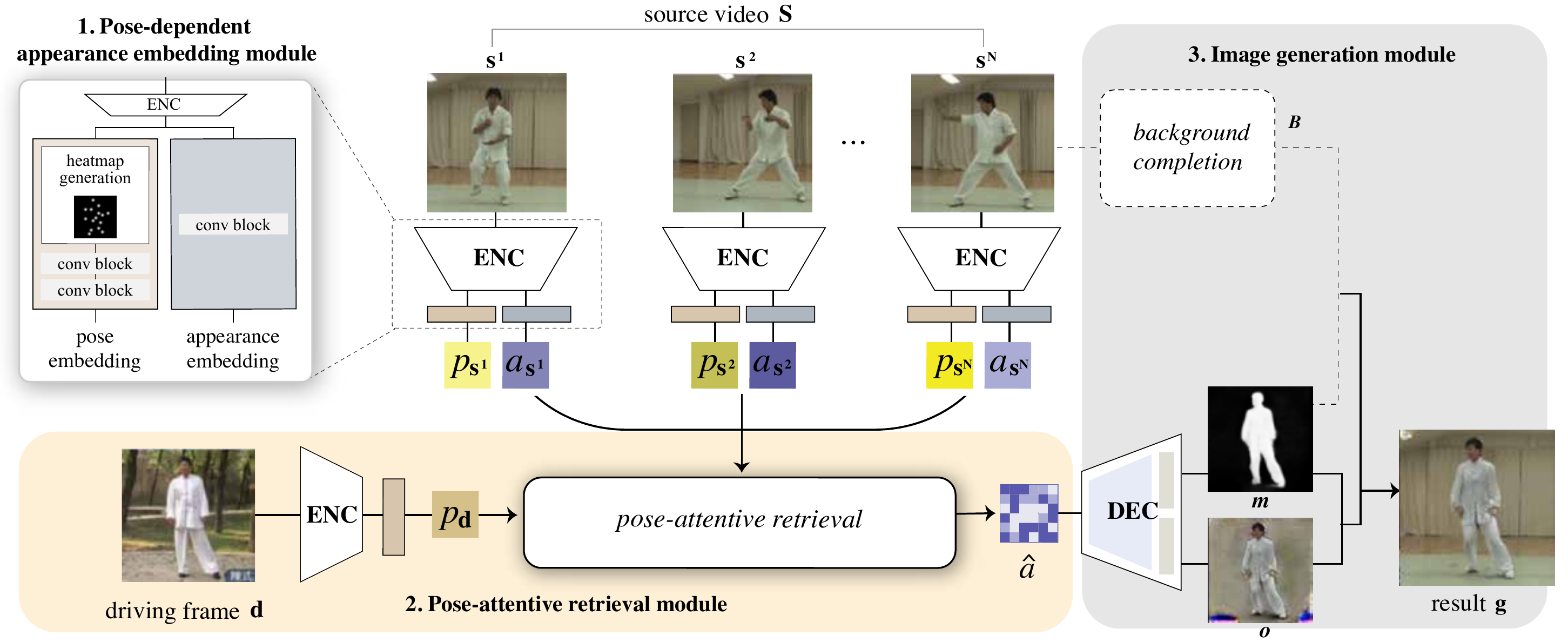}
    \caption{Overview of our work. It consists of (1) pose-dependent appearance embedding module, (2) pose-attentive retrieval module, and (3) image generation module. The pose-dependent appearance embedding module takes frames in the source video to embed (pose, appearance) pairs. Pose attention block retrieves appearance correspondent to the driving pose. The driving pose is extracted using the shared pose embedding network. Finally, image generation module generates the results.}
    \label{fig:overview}
\end{figure}

\subsection{Overview}
Given a source video $\mathbf{S}=\{\mathbf{s}^n\}_{n=1,2...,N}$ and a driving video $\mathbf{D}=\{\mathbf{d}^t\}_{t=1,2...,T}$, our task is to synthesize a video where the motion of the foreground object is similar to that of $\mathbf{D}$ while the appearance is same as that of $\mathbf{S}$.
Note that we deal with both cases of the single source frame $N=1$ and the multiple source frames $N>1$.

The overall architecture is illustrated in~\fref{fig:overview}.
Our model consists of following modules: (1) pose-dependent appearance embedding module, (2) pose-attentive retrieval module, and (3) image generation module.
The pose-dependent appearance embedding module takes $S$ as input, and extracts the pose and the appearance representations for each frame in $S$ independently, using a shared embedding network.
The pose attentive retrieval module extracts the pose from the driving frame, and searches for proper appearance features in the source frames using our spatio-temporal attention mechanism.
Finally, our decoder synthesizes outputs using the appearance features and one of the source images for the background.

Our network is trained using a large number of videos containing subjects in the same object category. 
At each iteration, the network is trained using two videos with different identities by minimizing the self-reconstruction loss and the cross-identity reconstruction loss. 
After training, it generates output video frame  conditioned on $S$ and each frame $d^t$ in the driving video. 
In the following, we describe our method in detail.

\subsection{Network Architecture}
\noindent\textbf{Pose-dependent appearance embedding}
Given a source video $S$, we extract poses $P_{S}$ and their corresponding appearance feature $A_{S}$. 
Each $s^n$, $n$-th frame of $S$, is processed independently through the shared embedding network, as illustrated in the upper side of~\fref{fig:overview}. 
It starts by putting a single image $s^n$ into the shared encoder. 
The features extracted from the encoder, then, goes through two parallel streams, for the pose and the appearance representation, respectively. 
% The feature extracted from the encoder is delivered in two separated streams, for the pose and the appearance representation, respectively. 
In the pose stream, it first reduces the number of channels to $K$ with a convolutional layer. 
Here, $K$ implicitly determines the number of keypoints to be extracted by the network. 
Each channel is condensed to represent a key point in a spatial domain by using the method proposed in ~\cite{Jakab:2018:imm}. 
In order to capture a local relationship between keypoints, the extracted keypoints are processed by a couple of convolutional layers. 
In case of the appearance stream, the appearance representation is embedded through a convolutional layer from the encoder feature. 
Each $s^n$ gets a $(p_{s^n}, a_{s^n})$ pair, resulting in $P_{S}=\{p_{s^n}\}_{n=1,2...,N}$ and $A_{S}=\{a_{s^n}\}_{n=1,2...,N}$, which are concatenated in time-dimension respectively.  
%$P_{S}=\{p_{s^1},\cdots,p_{s^N}\}$ and $A_{S}=\{a_{s^1},\cdots,a_{s^N}\}$

% \begin{figure}[t]
% \begin{center}
% \includegraphics[width=1\linewidth]{figures/non_local_re.pdf}
% \end{center}
%   \caption{Pose attention appearance retrieval module. Here, $\bigotimes$ represents matrix multiplication.}
% \label{fig:non_local}
% \end{figure}

\noindent\textbf{Pose-attentive retrieval block}
Given $P^S$, $A^{S}$, and a driving frame $d$, the goal is to synthesize a figure with the same identity as in the source video but in the pose of the subject in the driving frame.
To this end, a computational block that matches poses between frames to retrieve appearance information for the synthesis is designed. 
We take insights from a attention mechanism in the previous memory networks~\cite{oh:2019:space-time,oh:2019:onion}. However, different from ~\cite{oh:2019:space-time,oh:2019:onion} that matches visual features, we explicitly design our module to match pose keypoints with a different motivation.  

To match between poses, the driving pose $p_{d}$ is first extracted through the same pose embedding network used for the source video.
Then, similarities between the source poses $P_S$ and the driving pose $p_{d}$ are computed to determine where to retrieve relevant appearance information from the source appearance $A_S$.
Specifically, soft weights are computed in a non-local manner by comparing every pixel locations in the source pose embedding $P_S$ with every pixels of the driving pose embedding $p_{d}$.
Finally, the appearance $\hat{a}$ correspondent to $p_{d}$ is reassembled by taking a weighted summation of the source appearance features $A_S$ with regard to the soft weights.
Our appearance retrieval operation can be summarized as:
\begin{equation}
    \hat{a}^{i} = \frac{1}{Z}\sum_{\forall{j}}f(p_{d}^{i}, P_S^{j})A_{S}^{j},
\end{equation}
where $Z = \sum_{\forall{j}}f(p_{d}^{i}, P_S^{j})$ is a normalization factor, and $i$ and $j$ indicate the pixel location of the driving and the source pose features, respectively. 
At first, it computes the attention map using the similarity function $f$,
The similarity function $f$ is defined as follow:
\begin{equation}
    f(p_{d}^{i}, P_S^{j})= \text{exp}({p_{d}^{i}\cdot P_S^{j}}),
\end{equation}
where $\cdot$ is dot-product.

\noindent\textbf{Image generation}
Contrary to the object parts that change according to the driving pose, the background should remain still regardless of the variation of poses. 
Thus, the image generation module should focus on generating the object parts while the background are directly obtained from the source images.
To this end, the decoder produces two outputs by dividing the output branches; one to create the object parts $o$ and the other to create a mask for the object $m$.
% In case of the object part, it generates object in novel pose from the retrieved feature using simple decoder. 
% The last part of the decoder is divided into two branches, one to create the object part $O$ and the other to create a mask $M$ for the object. 
$m$ is a single channel probability map obtained by applying the sigmoid function. 
% We take different approach to get the background $B$ depending on the property of the dataset.
% If videos in the dataset have the dynamic backgrounds, one of source images is directly used for the background $B$.
% In other case, where videos have a static background, one of source image is 
We compute the background image $B$ by feeding one of source images to a simple encoder-decoder network to inpaint the occluded part.
Formally, we get the result by
\begin{equation}
    g=m \odot o +(1-m) \odot B,
\end{equation}
where $\odot$ represents element-wise multiplication.

\begin{figure}[t]
\begin{center}
\includegraphics[width=\linewidth]{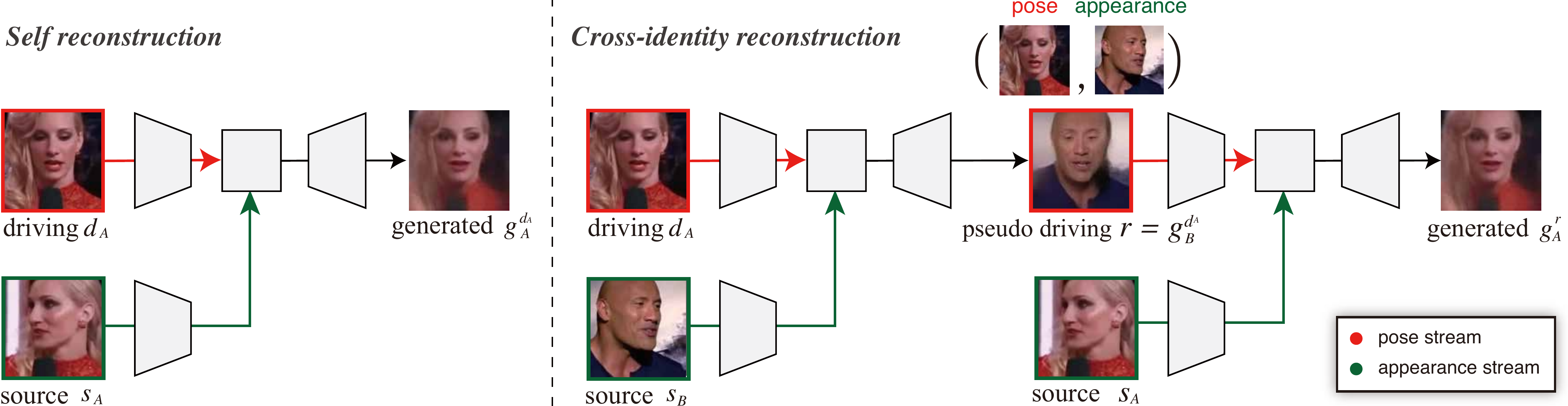}
\end{center}
   \caption{Self-reconstruction and cross-identity reconstruction training scheme.}
\label{fig:loss}
\end{figure}

\subsection{Training and Losses}
Our generator is trained using two main losses: self-reconstruction loss and cross-identity reconstruction loss.
The distinction between them depends on whether they use the same identity of objects on the source and the driving images, as illustrated in~\fref{fig:loss}.

\noindent\textbf{Self-reconstruction loss}
If source images $s_A$ and the driving image $d_A$ have the same identity $A$, the generated image $g_A^{d_A}$ has the identity of $A$ and the pose from $d_A$, which is equal to $d_A$. 
Therefore the network can be trained by using $d_A$ as the ground-truth image of $g_A^{d_A}$ in a self-supervised way. 
Note that we split a single video of an identity into source images $s_{identity}$ and driving image $d_{identity}$, and there is no overlap between them.

We impose the reconstruction loss on the pair of $(g_A^{d_A},d_A)$ to train the overall network. 
Additionally, we add adversarial loss for the quality of generated image. Thus the self-reconstruction loss is formulated as follows:
\begin{equation}
    \begin{gathered}
        L_{self}=L_{GAN}(G,D)
        +\lambda_{FM}L_{FM}(G,D)\\
        +\lambda_{VGG}L_{VGG}(g_A^{d_A},d_A).
    \end{gathered}
\end{equation}
Here, we give adversarial loss $L_{GAN}(G,D)$ using the least-square GAN formulation~\cite{mao2017least}.
$L_{FM}(G,D)$ represents the discriminator feature-matching loss ~\cite{wang2018high} and $L_{VGG}(g_A^{d_A},d_A)$ is the perceptual loss using the pretrained VGG19 network.

This objective is commonly used to train  previous motion transfer networks in a self-supervised way~\cite{Wiles:2018:X2Face,Siarohin:2018:monkeynet}.
However, using this self-reconstruction alone can possibly cause severe issues in the real testing, where subjects in the source and the driving image are different.
For example, only working with samples from the same video, the model can simply learn to assume the source and the driving subjects to have the same identity.
Thus the model may not be prepared for dealing with subjects that show a significant appearance difference. 
We address this issue by additionally imposing a novel cross-identity reconstruction loss that simulates the real testing cases, where the identity of subjects in the source and the driving are different.

\noindent\textbf{Cross-identity reconstruction loss}
Consider transferring motion between different identities, which is our real task. 
To train a network in this scenario, it requires a lot of pairs of different subjects in the same posture. 
However, obtaining such pairs of images is not only time-consuming, but also results in annotations that involve subjective judgments.
% As our self reconstruction loss is based on the pixel similarity between frames in the same video, it is difficult to train our task using the pairs of different videos.
% 
We overcome the challenge by two consecutive generation procedures between different subjects.
The key idea behind our method is that we synthesize an image that has a different identity but the same pose as the one in the source video, then we use it as a pseudo-driving frame to transform an arbitrary frame in the source. 
In this case, the output image should be the same as the original source frame with the same pose since the poses of both the driving and the source image are identical.
By minimizing the pixel distance, the network is enforced to preserve the identity-relevant content in the output.
Although our method shares a similar concept with existing self-supervision methods~\cite{Bansal:2018:RecycleGAN,zhu:2017:unpaired}, our method presents a novel approach for tackling the cross-identity motion transfer in that it provides a way of supervision using different identities through the pseudo-driving image.
% We transfer the pose of one subject to the other, and use the generated image as a new driving image. As a result, the original pose is reconstructed in the output image through two generation steps. 
% ~\fref{fig:loss} illustrates the overall procedure, which has a cycle form. 
% Note that our loss function is fundamentally different from the existing cycle loss in ~\cite{zhu:2017:unpaired}. 
% Our cycle reconstruction loss contributes to disentanglement of pose and appearance as well as image generation conditioned on them, while cycle loss in~\cite{zhu:2017:unpaired} is designed for domain translation.

\fref{fig:loss} illustrates the procedure of the cross-identity reconstruction scheme.
First, we synthesize a pseudo-driving frame using two different videos.
Specifically, we use $d_A$ and $s_B$ as the driving and the source image, respectively.
The generator synthesizes the output $r=g_B^{d_A}$, where the object $B$ mimics the pose in $d_A$. 
Then, we perform another generation using the output $r$ as the pseudo-driving image and $s_A$ as the source images.
In this case, the generator outputs $g_A^{r}$ that has the identity of $A$ and the pose of $r$.
Because the pose of $r$ is $d_A$, $g_A^{r}$ and $d_A$ can be expected to be identical. We minimize a reconstruction loss between this pair of images with an adversarial loss similar to the self-reconstruction:

\begin{equation}
    \begin{gathered}
        L_{cross}=L_{GAN}(G,D)
        +\lambda_{FM}L_{FM}(G,D)\\
        +\lambda_{VGG}L_{VGG}(g_A^{r},d_A).
    \end{gathered}
\end{equation}

\noindent\textbf{Full objective}
Our final training objective is described as
\begin{equation}
    \begin{gathered}
        L_{total}=L_{self}+L_{cross},%+\mathcal{L}_{other}
    \end{gathered}
\end{equation}
which is optimized by solving a minimax optimization problem on an adversarial learning framework.

\subsection{Inference}
At the inference time, our network takes a driving video and source images as input. 
Note that there is no limitation to the number of source images. 
At first, it performs the pose-dependent appearance embedding on all of the source images. 
Each source image is processed individually using the shared network. 
Regardless of the number of source frames used at the training time, our network can fully leverage diverse observations from all the source images.
The resulting image is generated based on a frame in the driving video using the pose-attentive retrieval module.
To transfer motions throughout the driving video, every frame in the driving video is forwarded one-by-one.

% \subsection{Implementation details}
% We implement our network using PyTorch. 
% We train our network using Adam optimizer with $\beta_1=0.5$, $\beta_2=0.99$. 
% The network is trained with a learning rate of 0.0002 for 100,000 iterations and we linearly decay the learning rate to 0 over the rest of iterations.
% We use 4 GPUs of Geforce RTX 2080 TI for training, and the training took about 4 days.
% At each training iteration, two videos contain different identities randomly are sampled. One is used for extracting source images, and the other is used as the driving video. Our network extracts 30 and 60 keypoints in the pose emebdding for Tai-chi dataset and VoxCeleb2 dataset respectively.

\section{Experiments}
\subsection{Experimental setup}
\noindent\textbf{Datasets}
~We conduct experiments on diverse object categories using following datasets. 
\textbf{VoxCeleb2}~\cite{Chung:2018:VoxCeleb2} dataset is a large dataset of human speaking videos downloaded from YouTube. VoxCeleb2 contains 150,480 videos of 6,112 celebrities with 145,569/ 4,911 train/test split.
\textbf{Thai-Chi-HD}~\cite{siarohin:2019:Taichi} contains 3,334 videos of Thai-chi performers. Among them, we use 3,049 videos for training and 285 videos for testing, following the official split. 
\textbf{BAIR robot pushing dataset~\cite{ebert:2017:bair}} is a collection of videos about robot arms moving and pushing various objects on the table. It provides 42,880 training and 128 test videos.
Note that the motion contained in the 3 datasets is different from each other, which is face, body, and machine, respectively. By using them, we show the effectiveness of our method on a variety of objects.

\noindent\textbf{Baselines}
~We compare our network with unsupervised motion transfer methods, which are able to expand on arbitrary objects.
Particularly for VoxCeleb2, we additionally compare our methods with multi-source motion transfer methods that use facial landmarks detected from a external landmark detector. To the best of our knowledge, we are not aware of any multi-source motion transfer methods designed for arbitrary objects. All of our baselines are listed below:

\begin{itemize}
    \item[--]X2Face~\cite{Wiles:2018:X2Face}: It transfers motion by warping objects (especially face) in the direction of $x$ and $y$ axis. Also, they utilize multiple source images by averaging all of the appearance features.
    \item[--]Monkey-net~\cite{Siarohin:2018:monkeynet}: It uses keypoints learned in self-supervised way to extract motion from general object categories. It transfers motion by estimating optical flows between keypoints.    \item[--]First-order~\cite{siarohin:2019:Taichi}: To express more complex and large motion, it uses keypoints and affine transformation for each keypoint.
    \item[--]NeuralHead~\cite{zakharov:2019:fews}: It transfers the appearance of source frames into a set of face landmarks. It fine-tunes subject-specific generator parameters using few shots of source images at test time.
    % By initializing subject-specific generator parameters using few shots of source images at test time, it transfers the motion on unseen subjects.
\end{itemize}

\begin{figure}[t]
    \centering
    \includegraphics[width=\linewidth]{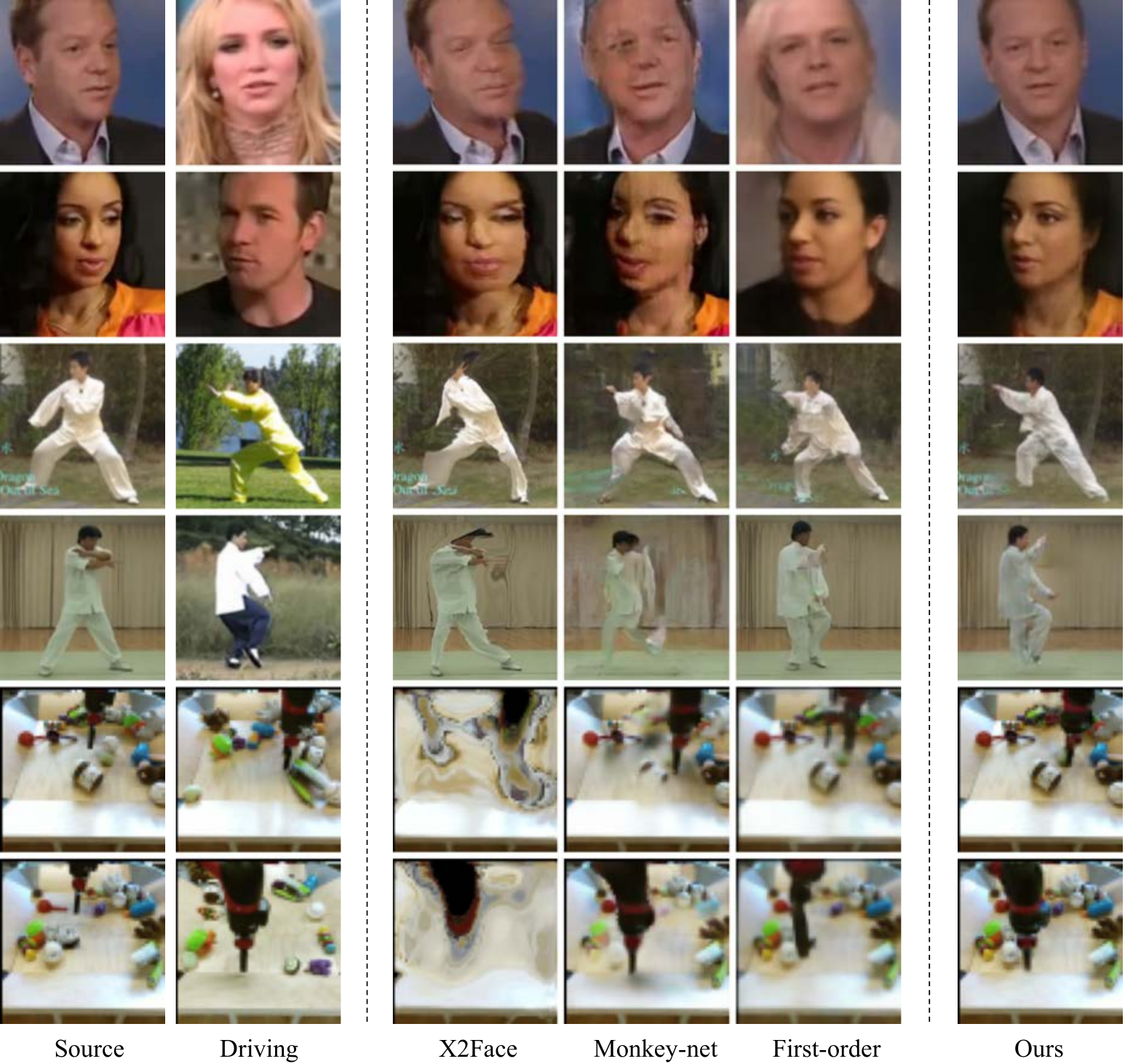}
    \caption{Qualitative results of single source animation. We conduct experiments on VoxCeleb2~\cite{Chung:2018:VoxCeleb2}, Thai-Chi-HD~\cite{siarohin:2019:Taichi}, and BAIR~\cite{ebert:2017:bair} datasets. Source images and driving images represent appearance and pose to condition on, respectively.}
    \label{fig:single}
\end{figure}

\noindent\textbf{Metrics}
~We evaluate motion transfer methods with the following criteria. We follow metrics proposed in Monkey-net~\cite{Siarohin:2018:monkeynet} and extend the evaluation scope by applying them on self reconstruction results and the motion transfer on different identity. 

\begin{itemize}
    \item[--]Fréchet Inception Distance (FID)~\cite{Heusel:2017:FID}: This score indicates the quality of generated images. It is computed by comparing the feature statistics of generated and real images to estimate how realistic the generated images are.
    \item[--]Average Keypoint Distance (AKD): It represents the average distance between keypoints of generated and driving videos, and evaluates whether the motion of driving video is transferred or not. We extract face and body keypoints using external keypoint detection networks~\cite{bulat:2017:face_landmarks,cao:2017:openpose}. 
    % However, this metric cannot be applied on other motion transfers because keypoints encompass not only the pose but also the identity-specific shape.
    \item[--]Average Euclidean Distance (AED): It shows the degree of identity preservation of generated images. In case of face and human body, we exploit external face recognition network~\cite{amos:2016:openface,hermans:2017:re-id} to embed identity-related features and compute average euclidean distance between them.
\end{itemize}

% \begin{table}[t]
% \begin{center}
% \caption{Comparison with other method on various datasets}
% \begin{tabular}{l|ccc|cc|ccc|cc|cc}
%           & \multicolumn{5}{c|}{VoxCeleb2}                         & \multicolumn{5}{c|}{Thai-Chi-HD}                           & \multicolumn{2}{c}{BAIR}          \\ \cline{2-13} 
%           & \multicolumn{3}{c|}{self} & \multicolumn{2}{c|}{other} & \multicolumn{3}{c|}{self} & \multicolumn{2}{c|}{other} & \multicolumn{1}{c|}{self} & other \\ \cline{2-13} 
%           & FID     & AKD     & AED   & FID    & AED         & FID     & AKD    & AED    & FID           & AED        & \multicolumn{1}{c|}{FID}   & FID   \\ \hline
% X2Face     & 48.63   & 22.23   & 0.58  & 55.46  & 0.69  & 94.99   & 4.88   & 1.33   & 101.74    & 1.42  & \multicolumn{1}{c|}{375.75}  & 379.60   \\
% Monkey & 27.98   & 18.26   & 0.46  & 52.49  & 0.66  & 30.43   & 2.85   & \textbf{0.93}   & 101.64    & 2.67  & \multicolumn{1}{c|}{24.28} & 75.57 \\
% First-order & 35.39& 2.52& 0.37 & 78.21 & 0.78 & \textbf{18.55} & \textbf{1.47} & 0.99 & \textbf{37.38} & 2.15 & \multicolumn{1}{c|}{31.80} & 52.39 \\
% Ours       & \textbf{17.24}   & \textbf{1.90}    & \textbf{0.38}  & \textbf{36.39}  & \textbf{0.62}  & 35.60  & 2.74 & 1.31  & 48.33 & 2.02 & \multicolumn{1}{c|}{\textbf{20.75}} & \textbf{34.81}  
% \end{tabular}
% \end{center}
% \label{table:single}
% \end{table}

\begin{table}[t!]
\centering
\caption{Quantitative comparison on various datasets. For all metrics, smaller values are better. The red and blue color indicate the first and the second ranks, respectively.}
\begin{tabular}{l|cc|cc|cc|cc|cc}
            & \multicolumn{4}{c|}{VoxCeleb2}                         & \multicolumn{4}{c|}{Thai-Chi-HD}                       & \multicolumn{2}{c}{BAIR}             \\ \cline{2-11} 
            & \multicolumn{2}{c|}{self} & \multicolumn{2}{c|}{cross} & \multicolumn{2}{c|}{self} & \multicolumn{2}{c|}{cross} & \multicolumn{1}{c|}{self}   & cross  \\ \cline{2-11} 
            & FID         & AKD         & FID          & AED         & FID          & AKD        & FID           & AED        & \multicolumn{1}{c|}{FID}    & FID    \\ \hline
X2Face~\cite{Wiles:2018:X2Face}  & 48.63 & 22.23 & 55.46 & 0.69 & 94.99 & 4.88  & 101.74 & \textbf{\textcolor{red}{1.42}} & \multicolumn{1}{c|}{375.75} & 379.60 \\
Monkey-net~\cite{Siarohin:2018:monkeynet}  & \textbf{\textcolor{blue}{27.98}}       & 18.26       & \textbf{\textcolor{blue}{52.49}}        & \textbf{\textcolor{blue}{0.66}} & \textbf{\textcolor{blue}{30.43}} & 2.85       & 101.64        & 2.67       & \multicolumn{1}{c|}{\textbf{\textcolor{blue}{24.28}}}  & 75.57  \\
First-order~\cite{siarohin:2019:Taichi} & 35.39  & \textbf{\textcolor{blue}{2.52}}  & 78.21  & 0.78 & \textbf{\textcolor{red}{18.55}} & \textbf{\textcolor{red}{1.47}}  & \textbf{\textcolor{red}{37.38}} & 2.15  & \multicolumn{1}{c|}{31.80}  & \textbf{\textcolor{blue}{52.39}}  \\
Ours        & \textbf{\textcolor{red}{17.24}} & \textbf{\textcolor{red}{1.90}} & \textbf{\textcolor{red}{36.39}} & \textbf{\textcolor{red}{0.62}} & 35.60 & \textbf{\textcolor{blue}{2.74}} & \textbf{\textcolor{blue}{48.33}} & \textbf{\textcolor{blue}{2.02}} & \multicolumn{1}{c|}{\textbf{\textcolor{red}{20.75}}}  & \textbf{\textcolor{red}{34.81}} 
\end{tabular}
\label{table:single}
\end{table}

\subsection{Experimental results}
\noindent\textbf{Single image animation of arbitrary objects}
~As existing methods for arbitrary objects are not capable of taking multiple source images, we first compare those methods in the setting of single source image.
\Tref{table:single} shows the quantitative comparison on all datasets.
In VoxCeleb2 and BAIR, our network outperforms all baselines methods in all metrics.
In Thai-Chi-HD, however, our results are worse than those of First-order~\cite{siarohin:2019:Taichi} especially for self-reconstruction.
Although First-order improves the warping-based approach using a two-stage warping with hallucinating occluded background, its generalization is limited to only Thai-Chi-HD with a self-supervision setting.
On the other hand, our method performs well on all datasets in general.
\fref{fig:single} shows qualitative results.
As can be seen, our method shows the effectiveness in transferring various types of motion. 
Warping-based methods work for the small movement change, e.g. leg regions of the third row in~\fref{fig:single}.
However, they have difficulties in synthesizing complex scenes with challenging viewpoint changes.
% In addition, all baseline methods lose the original shape of face and table in VoxCeleb2 and BAIR, and First-order method copies the content from a driving image, which is undesirable. 
% In contrast, our results successfully maintain the identity while accurately changing the pose.

\noindent\textbf{Comparison with multi-source methods}
~We compare results of using multiple frames.
For fair comparison with NeuralHead~\cite{zakharov:2019:fews}, we use the test set and the results provided by the authors of NeuralHead~\cite{zakharov:2019:fews}.
%because there is no official code of the method.
The test set is composed of 3 subsets with 1, 8, and 32 source frames in a self-reconstruction setting.

\begin{figure}[t!]
    \centering
    \includegraphics[width=\linewidth]{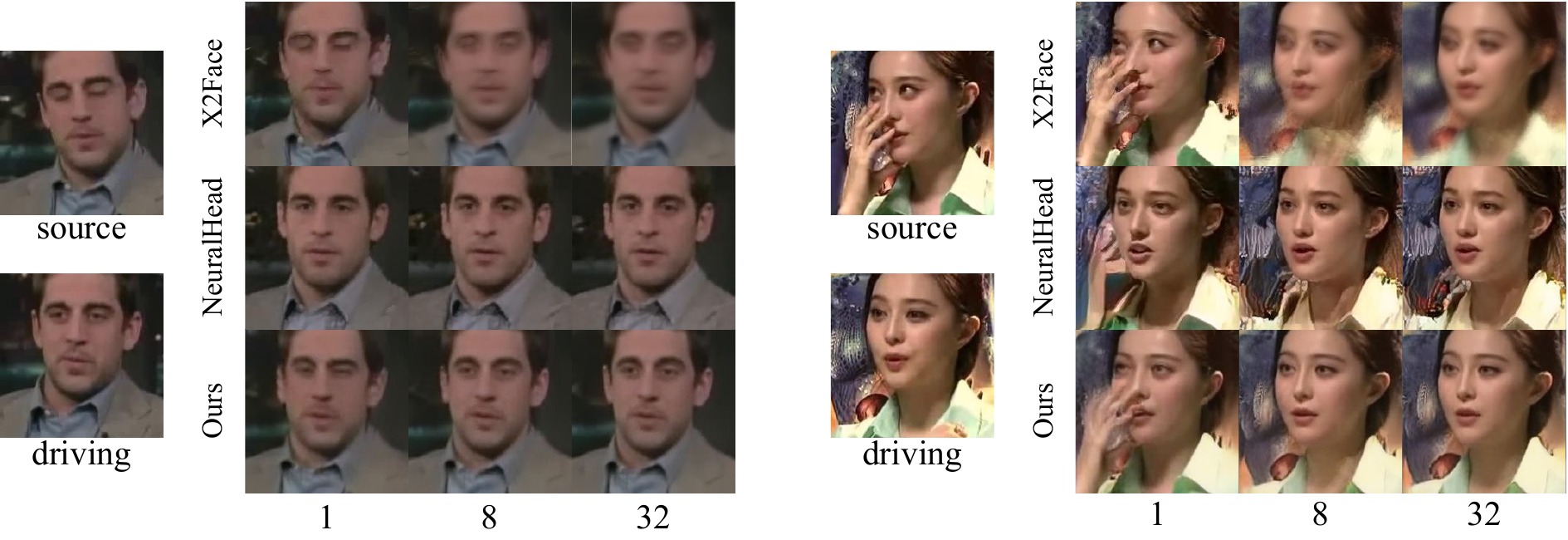}
    \caption{Comparison with face few-shot motion transfer methods. We compare results with varying number of source images.}
    \label{fig:few-shot}
\end{figure}

\Tref{table:multisources} represents the quantitative comparison. 
As can be seen, our method significantly outperforms X2Face and is competitive to NeuralHead.
This is because NeuralHead uses keypoints extracted from the pre-trained landmark detector and produces sharper images, while our method learns to estimate the pose without supervision.
% Our results come close to the AKD scores as more source frames are used. 
However, NeuralHead is specifically designed for facial videos with external landmarks, therefore the applicability to arbitrary objects is limited.
Note that our method outperforms both baselines on AED.
It indicates that our method better preserves the identity of the source due to our cross-identity training scheme, which is also observed in~\fref{fig:few-shot}.

\begin{table}[t!]
\centering
\begin{minipage}{.58\linewidth}
\caption{Quantitative results of few-shot motion transfer methods on VoxCeleb2.}
% \centering
\begin{adjustbox}{width=1\linewidth}
\begin{tabular}{c|ccc|ccc|ccc}
    & \multicolumn{3}{c|}{X2Face~\cite{Wiles:2018:X2Face}} & \multicolumn{3}{c|}{NeuralHead~\cite{zakharov:2019:fews}} & \multicolumn{3}{c}{Ours} \\ \cline{2-10} 
    & 1       & 8       & 32      & 1         & 8        & 32       & 1      & 8      & 32     \\ \hline
FID & 97.37   & 110.29  & 127.32  & 50.10 & 45.88 & \textbf{44.85} & 56.19  & 60.17  & 60.80  \\
AKD & 113.79  & 96.18   & 142.09  & 4.48 & 3.71 & \textbf{3.45}     & 7.47   & 4.26   & 4.09   \\
AED &   0.67  & 0.80  & 0.79 & 0.54     & 0.45     & 0.43    & 0.46  & 0.33  & \textbf{0.29} 
\end{tabular}
\end{adjustbox}
\label{table:multisources}
\end{minipage}
\hfill
\begin{minipage}{0.37\linewidth}
\caption{Ablation study on the cross-identity training scheme.}
% \centering
\begin{adjustbox}{width=1.0\linewidth}
\begin{tabular}{c|ccc|cc}
                   & \multicolumn{3}{c|}{self} & \multicolumn{2}{c}{cross} \\ \cline{2-6} 
                   & FID     & AKD    & AED    & FID          & AED        \\ \hline
w/o cross & 18.49   & 5.79   & 0.39   & 55.11        & 0.76       \\
w/ cross  & 17.24   & 1.90   & 0.38   & 36.39        & 0.62      
\end{tabular}
\end{adjustbox} 
\label{table:ablation}
\end{minipage}
\end{table}

\subsection{Human evaluation}
We conducted a user study using Amazon Mechanical Turk (AMT) to compare our results with previous warping based motion transfer methods~\cite{Wiles:2018:X2Face,Siarohin:2018:monkeynet,siarohin:2019:Taichi} subjectively. %X2Face~\cite{Wiles:2018:X2Face} and Monkey-net~\cite{Siarohin:2018:monkeynet}
We randomly selected 50 results of VoxCeleb2~\cite{Chung:2018:VoxCeleb2}, Thai-chi-HD~\cite{siarohin:2019:Taichi} and BAIR robot pushing~\cite{ebert:2017:bair} datasets. We used a single frame for source, and uniformly sampled 32 frames from the driving video. Source images and corresponding driving videos are matched randomly to make pairs of distinct identities.
In the user study, users ranked the result videos according to the following criteria: (1) whether the identity of the source image is preserved or not, and (2) whether the motion of the driving video is transferred or not. 
Each question was rated by 10 people and we averaged the total ranks. The results are shown in \Tref{table:user study}. 
As can be seen, our results are preferred over previous warping-based methods on VoxCeleb2 and BAIR with a large margin.
Even though our method achieves lower quantitative scores than First-order~\cite{siarohin:2019:Taichi} on Thai-Chi-HD in~\Tref{table:single}, the user study shows that our results are preferred by users as much as those of First-order.

\begin{table}[t!]
\centering
\begin{minipage}{.58\linewidth}
\caption{Human evaluation on AMT. The scores indicate average ranking values among 4 methods.}
\begin{adjustbox}{width=1\linewidth}
\begin{tabular}{l|cccc}
          & X2Face~\cite{Wiles:2018:X2Face} & Monkey-net~\cite{Siarohin:2018:monkeynet} & First-order~\cite{siarohin:2019:Taichi} &Ours \\ \hline
VoxCeleb2 & 3.116  & 2.884  & 2.414    & \textbf{1.586}  \\
Thai-Chi-HD   & 3.438  & 3.048 & 1.792   & \textbf{1.722} \\
BAIR & 3.58 & 2.104 & 2.48 & \textbf{1.836}
\end{tabular}
\end{adjustbox}
\label{table:user study}
\end{minipage}
\hfill
\begin{minipage}{0.37\linewidth}
\caption{Ablation study on pose-attentive retrieval module.}
\begin{adjustbox}{width=1\linewidth}
\begin{tabular}{r|cc|cc}
             & \multicolumn{2}{c|}{self} & \multicolumn{2}{c}{cross} \\ \cline{2-5} 
Method       & FID         & AKD         & FID          & AED        \\ \hline
ConcatAppearance       & 17.57       & 4.21        & 63.52        & 0.92       \\
ConcatNearest & 23.69       & 4.27        & 52.13        & 0.87       \\
ConcatAverage      & 51.35       & 103.78      & 52.67        & 1.35       \\ \hline
Ours         & \textbf{17.24}& \textbf{1.90}& \textbf{36.39} & \textbf{0.62}      
\end{tabular}
\end{adjustbox}
\label{table:attention ablation}
\end{minipage}
\end{table}

\subsection{Analysis}

\begin{figure}[t!]
\centering
\begin{minipage}{.53\linewidth}
% \begin{figure}[]
    \centering
    \includegraphics[width=\linewidth]{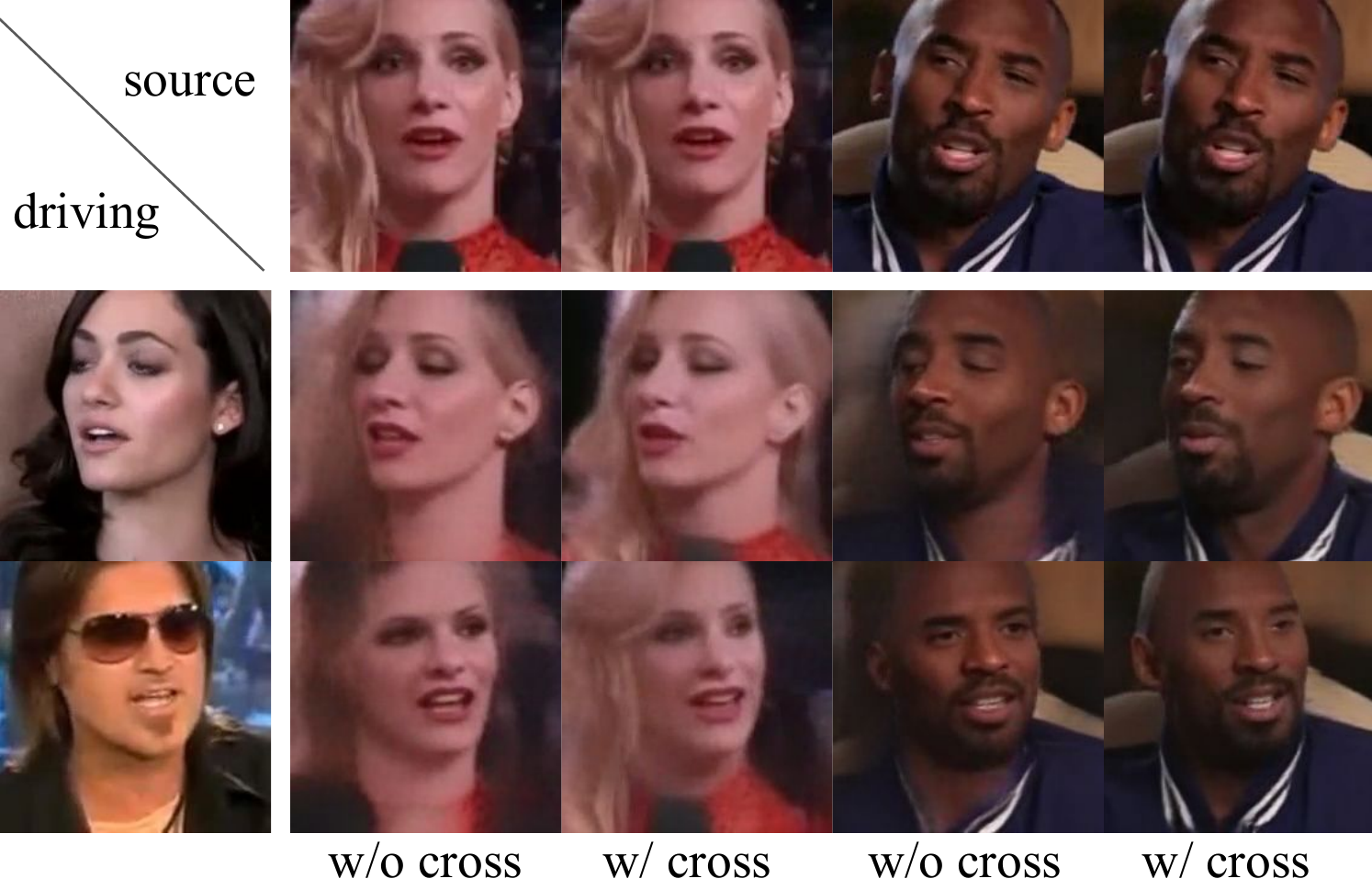}
    \caption{Ablation on the cross-identity reconstruction training scheme.}
    \label{fig:cycle}
% \end{figure}
\end{minipage}
\hfill
\begin{minipage}{.42\linewidth}
    \centering
    \includegraphics[width=\linewidth]{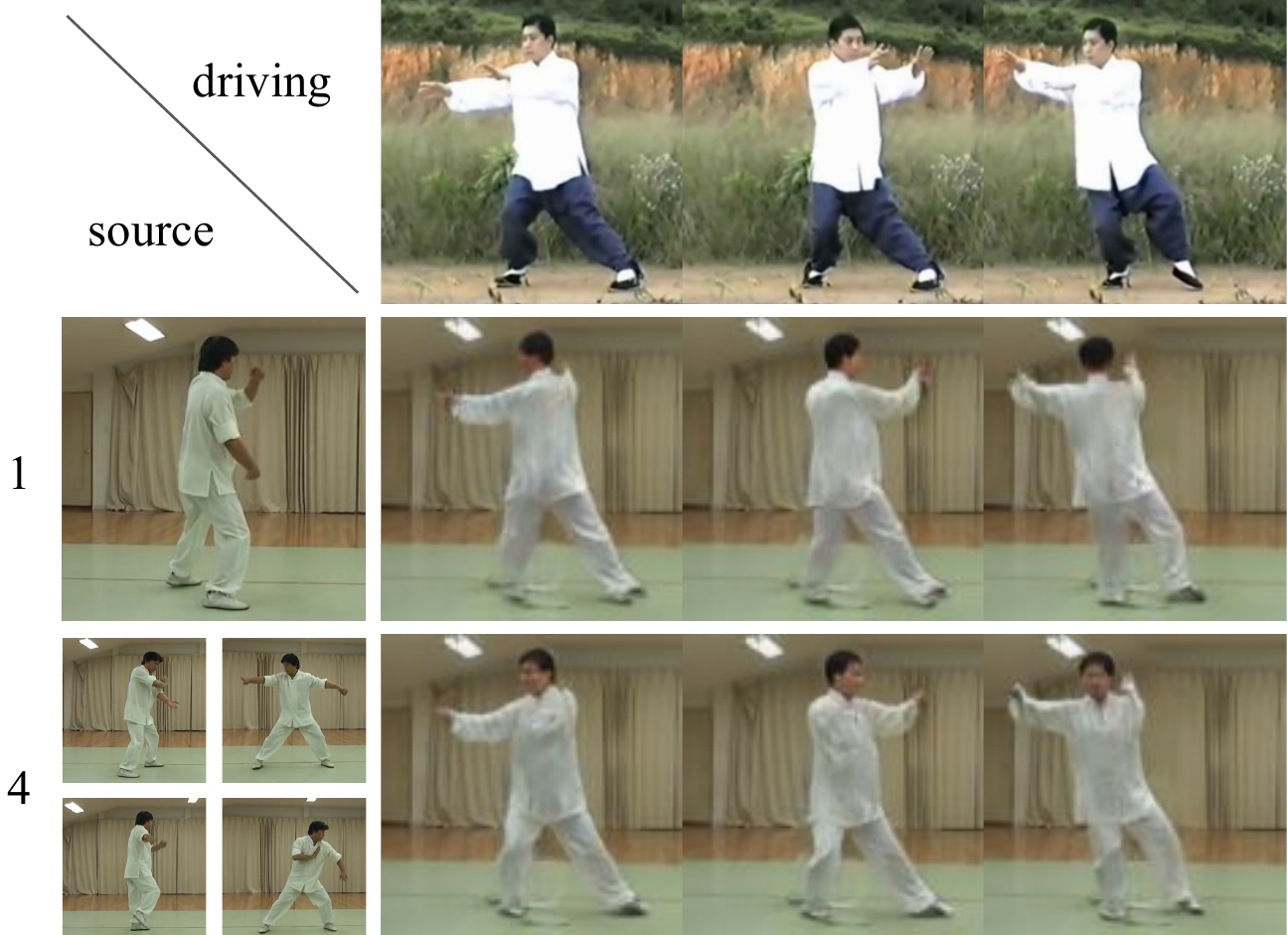}
    \caption{Ablation on using multiple source frames on Thai-Chi-HD.}
    \label{fig:multi_taichi}
\end{minipage}
% \vspace{-10pt}
\end{figure}

\noindent\textbf{Efficacy of cross-identity reconstruction loss}
We conduct ablation study on the cross-identity training scheme using VoxCeleb2 dataset by transferring motion between people of different genders and ages. The quantitative and qualitative results are shown in~\Tref{table:ablation} and~\fref{fig:cycle}.
As represented in the first row of each example in~\fref{fig:cycle}, the self-reconstruction training can disentangle shape and appearance. However, it is not able to further disentangle identity-specific shape such as face shape or skeleton. For example, results without the cross-identity reconstruction training show the tendency that the identity of source images is not preserved for all the driving videos. On the other hand, the identity is well preserved across diverse driving videos with the cross-identity training scheme. 

Moreover, the results using only the self-reconstruction training scheme show implausible output in regions surrounding the face such as hair, neck, and shoulder.
We attribute it to the fact that the self-reconstruction training scheme focuses on the facial region, which is the most different part between frames in the same video. 
With cross-identity training, our network is enforced to consider all parts of a person, thereby further improves the quality.
The efficacy of our cross-identity reconstruction training scheme is also demonstrated in~\Tref{table:ablation}. When adding our cross-identity reconstruction, both image quality and identity preservation are improved.

\noindent\textbf{Efficacy of pose-attentive retrieval module}
For a deeper analysis on pose-attentive retrieval module, we replace the module with following baselines.
\textbf{ConcatAppearance} concatenates pose and appearance features. It can take only a single source image as input.
\textbf{ConcatNearest} takes multiple source images as input and then selects a single image that has the most similar pose with the driving pose using Euclidean distance of the keypoints. 
\textbf{ConcatAverage} takes the average of feature maps of multiple images.
As can be seen in ~\Tref{table:attention ablation}, ConcatAppearance lowers the quality, especially in AKD. ConcatNearest could not improve the quality, as well.
Moreover, ConcatAverage deteriorates performance.

% \begin{table}[h]
% \centering
% \caption{Ablation study on pose-attentive retrieval module.}
% \begin{tabular}{r|cc|cc}
%              & \multicolumn{2}{c|}{self} & \multicolumn{2}{c}{cross} \\ \cline{2-5} 
% Method       & FID         & AKD         & FID          & AED        \\ \hline
% Concat       & 17.57       & 4.21        & 63.52        & 0.92       \\
% Nearest Pose & 23.69       & 4.27        & 52.13        & 0.87       \\
% Average      & 51.35       & 103.78      & 52.67        & 1.35       \\ \hline
% Ours         & \textbf{17.24}& \textbf{1.90}& \textbf{36.39} & \textbf{0.62}      
% \end{tabular}
% \label{table:attention ablation}
% \end{table}

% \begin{itemize}
%     \item[--]Concat: We concatenate pose and appearance features
% \end{itemize}

\noindent\textbf{Using multiple source images at inference time}
~\fref{fig:multi_taichi} demonstrates the effectiveness of utilizing multiple source images.
The driving images are shown on the first row, and the second and the third row show the output images when 1 or 4 source images are given, respectively.
With a single source image, reconstructing occluded parts is difficult as shown in the head region in the second row.
On the other hand, exploiting multiple source images alleviates the issue by copying and pasting the occluded region from the source images.
The performance gain is also shown quantitatively in~\Tref{table:multisources}.

\noindent\textbf{Visualization}
\begin{figure}[t!]
    \centering
    \includegraphics[width=\linewidth]{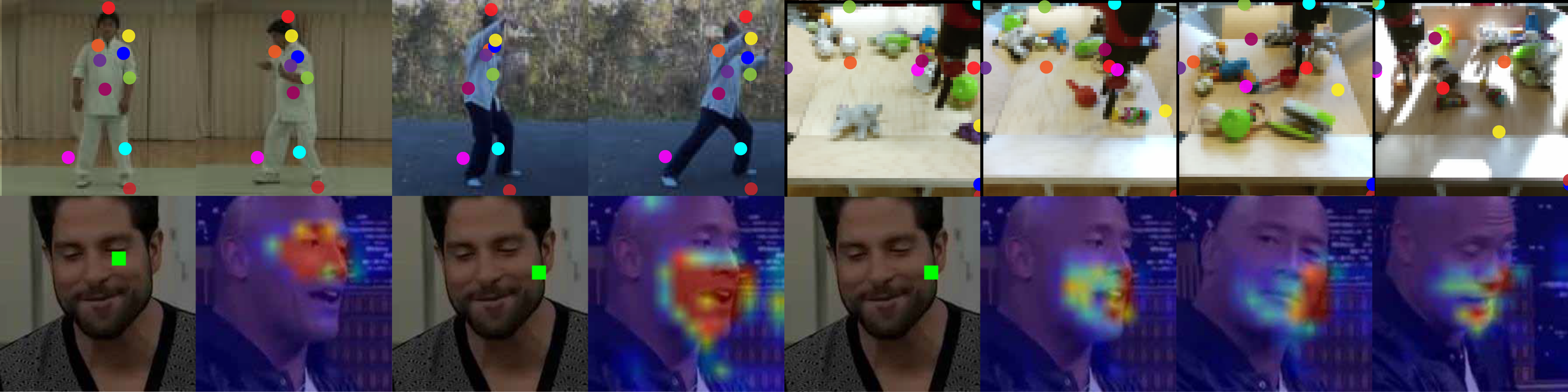}
    \caption{Visualization of keypoints and attention. We visualize learned keypoints on Thai-Chi-HD and BAIR datasets in the first row. The second row represents activated attention maps according to driving images. }
    \label{fig:visualization}
    % \vspace{-10pt}
\end{figure}
To understand the learned representation of our method, we visualize the learned keypoints and attention in ~\fref{fig:visualization}.
The first row illustrates 10 keypoints extracted from the given frames.
% We selectively visualize 10 keypoints around objects for clarification.  
In the second row, we visualize the attention map in the source images corresponding to the query points in the driving pose.
% As can be seen, our method effectively learns keypoints of arbitrary objects without annotation, and spatio-temporal attention using the learned keypoints.

\section{Conclusion}
In this paper, we presented a pose-attentive video reassembling method to tackle the problem of cross-identity motion transfer method for arbitrary objects. 
% Previous warping based approaches were limited in terms of handling large motion and occlusion. 
Unlike existing warping-based approaches, our pose-attentive reassembling method better handles a wide range of motion and is also able to exploit multiple sources images to robustly reconstruct occluded parts.
Furthermore, our cross-identity training scheme disentangles identity-relevant content from object pose to preserve identity in the setting of cross-identity motion transfer.
% Experimental results show that our method outperforms existing methods in most cases on VoxCeleb2, Thai-Chi-HD, and BAIR datasets.

\noindent\textbf{Acknowledgement}
This work was conducted by Center for Applied Research in Artificial Intelligence
(CARAI) grant funded by DAPA and ADD (UD190031RD).
% \input{rebuttal.tex}

% ---- Bibliography ----
%
% BibTeX users should specify bibliography style 'splncs04'.
% References will then be sorted and formatted in the correct style.
%
\bibliographystyle{./splncs04.bst}
\bibliography{./egbib.bib}

\begin{thebibliography}{10}
\providecommand{\url}[1]{\texttt{#1}}
\providecommand{\urlprefix}{URL }
\providecommand{\doi}[1]{https://doi.org/#1}

\bibitem{amos:2016:openface}
Amos, B., Ludwiczuk, B., Satyanarayanan, M., et~al.: Openface: A
  general-purpose face recognition library with mobile applications. CMU School
  of Computer Science  \textbf{6} (2016)

\bibitem{balakrishnan:2018:unseen}
Balakrishnan, G., Zhao, A., Dalca, A.V., Durand, F., Guttag, J.: Synthesizing
  images of humans in unseen poses. In: Proceedings of the IEEE Conference on
  Computer Vision and Pattern Recognition. pp. 8340--8348 (2018)

\bibitem{Bansal:2018:RecycleGAN}
Bansal, A., Ma, S., Ramanan, D., Sheikh, Y.: Recycle-gan: Unsupervised video
  retargeting. In: Proceedings of the European Conference on Computer Vision
  (ECCV). pp. 119--135 (2018)

\bibitem{bulat:2017:face_landmarks}
Bulat, A., Tzimiropoulos, G.: How far are we from solving the 2d \& 3d face
  alignment problem?(and a dataset of 230,000 3d facial landmarks). In:
  Proceedings of the IEEE International Conference on Computer Vision. pp.
  1021--1030 (2017)

\bibitem{cao:2017:openpose}
Cao, Z., Simon, T., Wei, S.E., Sheikh, Y.: Realtime multi-person 2d pose
  estimation using part affinity fields. In: Proceedings of the IEEE Conference
  on Computer Vision and Pattern Recognition. pp. 7291--7299 (2017)

\bibitem{chan:2019:everybody}
Chan, C., Ginosar, S., Zhou, T., Efros, A.A.: Everybody dance now. In:
  Proceedings of the IEEE International Conference on Computer Vision. pp.
  5933--5942 (2019)

\bibitem{Chung:2018:VoxCeleb2}
Chung, J.S., Nagrani, A., Zisserman, A.: Voxceleb2: Deep speaker recognition.
  Proc. Interspeech 2018 pp. 1086--1090 (2018)

\bibitem{denton:2018:stochastic_prediction}
Denton, E., Fergus, R.: Stochastic video generation with a learned prior. arXiv
  preprint arXiv:1802.07687  (2018)

\bibitem{denton:2017:disentangledVideo}
Denton, E.L., et~al.: Unsupervised learning of disentangled representations
  from video. In: Advances in neural information processing systems. pp.
  4414--4423 (2017)

\bibitem{ding:2018:exprgan}
Ding, H., Sricharan, K., Chellappa, R.: Exprgan: Facial expression editing with
  controllable expression intensity. In: Thirty-Second AAAI Conference on
  Artificial Intelligence (2018)

\bibitem{ebert:2017:bair}
Ebert, F., Finn, C., Lee, A.X., Levine, S.: Self-supervised visual planning
  with temporal skip connections. arXiv preprint arXiv:1710.05268  (2017)

\bibitem{Goodfellow:2014:GAN}
Goodfellow, I., Pouget-Abadie, J., Mirza, M., Xu, B., Warde-Farley, D., Ozair,
  S., Courville, A., Bengio, Y.: Generative adversarial nets. In: Advances in
  neural information processing systems. pp. 2672--2680 (2014)

\bibitem{Ha:2020:marionnette}
Ha, S., Kersner, M., Kim, B., Seo, S., Kim, D.: Marionette: Few-shot face
  reenactment preserving identity of unseen targets. In: Proceedings of the
  AAAI Conference on Artificial Intelligence (2020)

\bibitem{hermans:2017:re-id}
Hermans, A., Beyer, L., Leibe, B.: In defense of the triplet loss for person
  re-identification. arXiv preprint arXiv:1703.07737  (2017)

\bibitem{Heusel:2017:FID}
Heusel, M., Ramsauer, H., Unterthiner, T., Nessler, B., Hochreiter, S.: Gans
  trained by a two time-scale update rule converge to a local nash equilibrium.
  In: Advances in Neural Information Processing Systems. pp. 6626--6637 (2017)

\bibitem{Jakab:2018:imm}
Jakab, T., Gupta, A., Bilen, H., Vedaldi, A.: Unsupervised learning of object
  landmarks through conditional image generation. In: Bengio, S., Wallach, H.,
  Larochelle, H., Grauman, K., Cesa-Bianchi, N., Garnett, R. (eds.) Advances in
  Neural Information Processing Systems 31, pp. 4016--4027. Curran Associates,
  Inc. (2018),
  \url{http://papers.nips.cc/paper/7657-unsupervised-learning-of-object-landmarks-through-conditional-image-generation.pdf}

\bibitem{kulkarni:2015:invgraphics}
Kulkarni, T.D., Whitney, W.F., Kohli, P., Tenenbaum, J.: Deep convolutional
  inverse graphics network. In: Advances in neural information processing
  systems. pp. 2539--2547 (2015)

\bibitem{lathuiliere:2020:attention}
Lathuili{\`e}re, S., Sangineto, E., Siarohin, A., Sebe, N.: Attention-based
  fusion for multi-source human image generation. In: The IEEE Winter
  Conference on Applications of Computer Vision. pp. 439--448 (2020)

\bibitem{ma:2017:PoseGuided}
Ma, L., Jia, X., Sun, Q., Schiele, B., Tuytelaars, T., Van~Gool, L.: Pose
  guided person image generation. In: Advances in Neural Information Processing
  Systems. pp. 406--416 (2017)

\bibitem{ma:2018:desentangledPerson}
Ma, L., Sun, Q., Georgoulis, S., Van~Gool, L., Schiele, B., Fritz, M.:
  Disentangled person image generation. In: Proceedings of the IEEE Conference
  on Computer Vision and Pattern Recognition. pp. 99--108 (2018)

\bibitem{mao2017least}
Mao, X., Li, Q., Xie, H., Lau, R.Y., Wang, Z., Paul~Smolley, S.: Least squares
  generative adversarial networks. In: Proceedings of the IEEE International
  Conference on Computer Vision. pp. 2794--2802 (2017)

\bibitem{oh:2019:space-time}
Oh, S.W., Lee, J.Y., Xu, N., Kim, S.J.: Video object segmentation using
  space-time memory networks. In: Proceedings of the IEEE International
  Conference on Computer Vision. pp. 9226--9235 (2019)

\bibitem{oh:2019:onion}
Oh, S.W., Lee, S., Lee, J.Y., Kim, S.J.: Onion-peel networks for deep video
  completion. In: Proceedings of the IEEE International Conference on Computer
  Vision. pp. 4403--4412 (2019)

\bibitem{saito:2017:TGAN}
Saito, M., Matsumoto, E., Saito, S.: Temporal generative adversarial nets with
  singular value clipping. In: Proceedings of the IEEE International Conference
  on Computer Vision. pp. 2830--2839 (2017)

\bibitem{Siarohin:2018:monkeynet}
Siarohin, A., Lathuili{\`e}re, S., Tulyakov, S., Ricci, E., Sebe, N.: Animating
  arbitrary objects via deep motion transfer. In: Proceedings of the IEEE
  Conference on Computer Vision and Pattern Recognition. pp. 2377--2386 (2019)

\bibitem{siarohin:2019:Taichi}
Siarohin, A., Lathuili\`{e}re, S., Tulyakov, S., Ricci, E., Sebe, N.: First
  order motion model for image animation. In: Advances in Neural Information
  Processing Systems 32, pp. 7137--7147. Curran Associates, Inc. (2019),
  \url{http://papers.nips.cc/paper/8935-first-order-motion-model-for-image-animation.pdf}

\bibitem{siarohin:2018:deformableHuman}
Siarohin, A., Sangineto, E., Lathuili{\`e}re, S., Sebe, N.: Deformable gans for
  pose-based human image generation. In: Proceedings of the IEEE Conference on
  Computer Vision and Pattern Recognition. pp. 3408--3416 (2018)

\bibitem{tran:2017:disentangledface}
Tran, L., Yin, X., Liu, X.: Disentangled representation learning gan for
  pose-invariant face recognition. In: Proceedings of the IEEE Conference on
  Computer Vision and Pattern Recognition. pp. 1415--1424 (2017)

\bibitem{Tulyakov:2018:MoCoGAN}
Tulyakov, S., Liu, M.Y., Yang, X., Kautz, J.: Mocogan: Decomposing motion and
  content for video generation. In: Proceedings of the IEEE conference on
  computer vision and pattern recognition. pp. 1526--1535 (2018)

\bibitem{villegas:2017:prediction}
Villegas, R., Yang, J., Zou, Y., Sohn, S., Lin, X., Lee, H.: Learning to
  generate long-term future via hierarchical prediction. In: Proceedings of the
  34th International Conference on Machine Learning-Volume 70. pp. 3560--3569.
  JMLR. org (2017)

\bibitem{vondrick:2016:VGAN}
Vondrick, C., Pirsiavash, H., Torralba, A.: Generating videos with scene
  dynamics. In: Advances In Neural Information Processing Systems. pp. 613--621
  (2016)

\bibitem{wang:2019:few-vid2vid}
Wang, T.C., Liu, M.Y., Tao, A., Liu, G., Catanzaro, B., Kautz, J.: Few-shot
  video-to-video synthesis. In: Advances in Neural Information Processing
  Systems. pp. 5014--5025 (2019)

\bibitem{wang2018high}
Wang, T.C., Liu, M.Y., Zhu, J.Y., Tao, A., Kautz, J., Catanzaro, B.:
  High-resolution image synthesis and semantic manipulation with conditional
  gans. In: Proceedings of the IEEE conference on computer vision and pattern
  recognition. pp. 8798--8807 (2018)

\bibitem{Wang:2018:vid2vid}
Wang, T.C., Liu, M.Y., Zhu, J.Y., Yakovenko, N., Tao, A., Kautz, J., Catanzaro,
  B.: Video-to-video synthesis. In: Advances in Neural Information Processing
  Systems. pp. 1152--1164 (2018)

\bibitem{Wiles:2018:X2Face}
Wiles, O., Sophia~Koepke, A., Zisserman, A.: X2face: A network for controlling
  face generation using images, audio, and pose codes. In: Proceedings of the
  European Conference on Computer Vision (ECCV). pp. 670--686 (2018)

\bibitem{zakharov:2019:fews}
Zakharov, E., Shysheya, A., Burkov, E., Lempitsky, V.: Few-shot adversarial
  learning of realistic neural talking head models. In: Proceedings of the IEEE
  International Conference on Computer Vision. pp. 9459--9468 (2019)

\bibitem{zhao:2018:motion_forecasting}
Zhao, L., Peng, X., Tian, Y., Kapadia, M., Metaxas, D.: Learning to forecast
  and refine residual motion for image-to-video generation. In: Proceedings of
  the European Conference on Computer Vision (ECCV). pp. 387--403 (2018)

\bibitem{zhu:2017:unpaired}
Zhu, J.Y., Park, T., Isola, P., Efros, A.A.: Unpaired image-to-image
  translation using cycle-consistent adversarial networks. In: Proceedings of
  the IEEE international conference on computer vision. pp. 2223--2232 (2017)

\end{thebibliography}
\end{document}